\documentclass{article}

\usepackage[dblblindworkshop, preprint]{neurips_2025}
\workshoptitle{TS4H: Learning from Time Series for Health}

\usepackage[utf8]{inputenc} 
\usepackage[T1]{fontenc}    
\usepackage{hyperref}       
\usepackage{url}            
\usepackage{booktabs}       
\usepackage{amsfonts}       
\usepackage{nicefrac}       
\usepackage{microtype}      
\usepackage{xcolor}         
\usepackage{amsmath}
\usepackage{natbib}
\usepackage{graphicx}
\usepackage{placeins}

\title{StableSleep: Source-Free Test-Time Adaptation for Sleep Staging with Lightweight Safety Rails}

\author{
  Hritik Arasu \\
  Department of Behavior and Brain Sciences\\
  University of Texas at Dallas\\
  Richardson, TX 75080 \\
  \texttt{hritik.arasu@UTDallas.edu} \\
  \And
  Faisal R. Jahangiri \\
  Department of Behavior and Brain Sciences\\
  University of Texas at Dallas\\
  Richardson, TX 75080 \\
  \texttt{faisal.jahangiri@utdallas.edu} \\
}

\begin{document}

\maketitle


\begin{abstract}
Sleep staging models often degrade when deployed on patients with unseen physiology or recording conditions. We propose a streaming, source-free test-time adaptation (TTA) recipe that combines entropy minimization (\emph{Tent}) with BatchNorm statistic refresh and two safety rails: an entropy gate to pause adaptation on uncertain windows and an EMA-based reset to reel back drift. On Sleep-EDF Expanded \citep{kemp2013sleepedf,sleepedfx2018,goldberger2000physionet}, using single-lead EEG (Fpz--Cz, 100~Hz, 30~s epochs; R\&K$\!\to\!$AASM mapping \citep{RechtschaffenKales1968,Iber2007AASM,Berry2012AASMUpdate,berry2015aasm}), we show consistent gains over a frozen baseline at seconds-level latency and minimal memory, reporting per-stage metrics and Cohen's $\kappa$ \citep{cohen1960kappa}. The method is model-agnostic, requires no source data or patient calibration, and is practical for on-device or bedside use.
\end{abstract}

\section{Introduction}
Deep models have markedly improved single-channel sleep staging, spanning convolutional and temporal pipelines (DeepSleepNet \citep{supratak2017deepsleepnet}, SeqSleepNet \citep{Phan2019SeqSleepNet}), fully convolutional segmentation (U-Time \citep{perslev2019utime}), high-rate cross-cohort models (U-Sleep \citep{Perslev2021USleep}), and attention-based variants (AttnSleep \citep{eldele2021attnsleep}). Still, models trained on one cohort degrade under real deployment shifts in montage, amplifiers, and population \citep{alvarezestevez2021interdb}. Centralizing new data or retraining per site is often misaligned with governance and privacy.

We focus on \emph{source-free} TTA: adapt a trained source model online during inference using only unlabeled target streams. Two ingredients are attractive for edge settings: refreshing BatchNorm (BN) statistics to track target distributions \citep{li2016adabn,Li2018AdaBN} and minimizing prediction entropy (Tent) while updating only normalization layers \citep{wang2021tent}. We pair them with two lightweight safeguards that we actually use in deployment-like streams: an \textbf{entropy gate} to suspend updates on low-confidence or artefactual windows, and an \textbf{EMA reset} to recover from drift. Related ideas include self-supervised test-time training (TTT) \citep{sun2020ttt}, source-free domain adaptation (SHOT) \citep{liang2020shot}, and continual/streaming TTA stability mechanisms (CoTTA) \citep{wang2022cotta}.

\paragraph{Contributions.}
(1) A simple, deployment-minded recipe (BN refresh $+$ Tent) with an entropy gate and EMA reset; (2) a reproducible single-lead (Fpz--Cz) evaluation on Sleep-EDF Expanded \citep{kemp2000microcontinuity,sleepedfx2018,goldberger2000physionet} with subject-disjoint splits and AASM-compliant mapping \citep{RechtschaffenKales1968,Iber2007AASM,Berry2012AASMUpdate,berry2015aasm}; (3) ablations of BN-only vs.\ Tent, gating, and resets, reported with accuracy, macro/weighted $F_1$, Cohen’s $\kappa$, balanced accuracy, MCC, and ECE.

\section{Background and Related Work}
\textbf{Sleep staging conventions and datasets.}
Clinical staging segments PSG into W/N1/N2/N3/REM using 30\,s epochs under R\&K and AASM \citep{RechtschaffenKales1968,Iber2007AASM,Berry2012AASMUpdate,berry2015aasm}. Sleep-EDF (Expanded) on PhysioNet provides Fpz--Cz and Pz--Oz EEG, EOG/EMG, and expert hypnograms \citep{kemp2000microcontinuity,kemp2013sleepedf,goldberger2000physionet,sleepedfx2018}. For broader external validation (future work here), MASS, SHHS, and ISRUC differ in hardware and cohorts \citep{oreilly2014mass,quan1997shhs,khalighi2016isruc} and are known to surface distribution shift.

\textbf{Supervised baselines and cross-cohort generalization.}
DeepSleepNet \citep{supratak2017deepsleepnet} and SeqSleepNet \citep{Phan2019SeqSleepNet} combine CNN features with temporal context; U-Time \citep{perslev2019utime} performs fully convolutional segmentation; U-Sleep \citep{Perslev2021USleep} scales to high sampling rates; AttnSleep \citep{eldele2021attnsleep} adds attention. Despite strong in-domain scores, inter-database evaluations report material drops across devices and cohorts \citep{alvarezestevez2021interdb}.

\textbf{Test-time and source-free adaptation.}
AdaBN aligns distributions via BN statistics \citep{li2016adabn,Li2018AdaBN}. Tent adapts by minimizing prediction entropy while updating only normalization layers \citep{wang2021tent}. TTT \citep{sun2020ttt} and SFDA (e.g., SHOT \citep{liang2020shot}) remove the need for source data at deployment. Continual/streaming TTA emphasizes stability (e.g., CoTTA \citep{wang2022cotta}). We adopt BN refresh $+$ Tent and add explicit gates/resets for streaming robustness, keeping compute and memory modest.

\section{Methods}
\label{sec:methods}
\subsection{Problem setting and streaming constraint}
Given a stream of $30$\,s epochs $\{x_t\}$ from single-lead EEG (Fpz--Cz), predict $y_t \in \{\text{W},\text{N1},\text{N2},\text{N3},\text{REM}\}$ online. Test labels are unavailable; we do not peek into future windows when adapting or post-processing.

\subsection{Dataset, preprocessing, and mapping}
We use Sleep-EDF Expanded with subject-disjoint train/val/test. Preprocessing: notch 50/60\,Hz, band-pass 0.3–45\,Hz, resample to 100\,Hz, segment into 30\,s epochs, and standardize per record using streaming running statistics (deployment-aligned). I/O and DSP use MNE-Python \citep{gramfort2013mne,gramfort2014mne}. We apply the standard R\&K$\!\to$AASM mapping \citep{RechtschaffenKales1968,Iber2007AASM,Berry2012AASMUpdate,berry2015aasm}.

\subsection{Architecture and source training}
The source model is a compact 1D CNN with depthwise-separable convolutions and squeeze-and-excitation (SE) blocks \citep{howard2017mobilenets,hu2018squeeze}; a lightweight temporal attention head summarizes features before the classifier. BN layers are explicit to support adaptation \citep{ioffe2015batch}. Training uses source subjects only with class-balanced focal loss \citep{lin2017focal}:
\[
\mathcal{L}_{\text{focal}} = -\sum_{c} \alpha_c (1-p_{t,c})^{\gamma} y_{t,c}\log p_{t,c},
\]
and prior-biased classifier initialization to stabilize early training \citep{menon2020longtail}. We use Adam (lr $10^{-3}$), warmup, and mild temporal/signal augmentations (jitter, amplitude scaling, short temporal masking).

\subsection{Test-time adaptation (Tent) with safety rails}
At deployment, only BN affine parameters and running statistics are updated by minimizing prediction entropy on streaming micro-batches $B$:
\[
\min_{\theta_{\text{BN}}}\ \mathcal{L}_{\text{ent}}(B)=\frac{1}{|B|}\sum_{x \in B}\Big(-\sum_{c} p_{\theta}(c|x)\log p_{\theta}(c|x)\Big),
\]
while the backbone remains frozen \citep{wang2021tent}. BN running means/variances are refreshed with momentum; we also evaluate a \emph{BN-only} baseline that recomputes BN statistics without gradients \citep{li2016adabn,Li2018AdaBN}. For stability, we (i) maintain an EMA of batch entropy and skip updates unless $\hat{H}_t\in[h_{\min},h_{\max}]$ (entropy gate), and (ii) keep an EMA snapshot of adapted BN parameters and reset to it when a drift criterion triggers (EMA reset), akin to continual TTA stabilizers \citep{wang2022cotta}. We apply a causal median filter (width 5) to reduce prediction flicker.

\subsection{Evaluation protocol and metrics}
Hyperparameters are selected on validation and reused unchanged on test; TTA never accesses labels. We report subject-wise means for accuracy, macro/weighted $F_1$, balanced accuracy, Cohen’s $\kappa$ \citep{cohen1960kappa}, MCC, and expected calibration error (ECE). For completeness, $\kappa=\frac{p_o-p_e}{1-p_e}$, where $p_o$ is observed agreement and $p_e$ is chance agreement. ECE is computed with standard binning: if $\mathcal{B}_m$ is bin $m$ with accuracy $\text{acc}(m)$ and average confidence $\text{conf}(m)$, then $\mathrm{ECE}=\sum_m \frac{|\mathcal{B}_m|}{N}\,|\text{acc}(m)-\text{conf}(m)|$ \citep{guo2017calibration}.

\section{Results}
\subsection{Aggregate performance}
We evaluate on validation and held-out test splits under single-lead EEG (Fpz--Cz). Aggregate metrics (mean across subjects) are shown in Table~\ref{tab:overall}. These summarize overall performance, agreement, class balance, correlation, and calibration.
\begin{table}[t]
\centering
\caption{Aggregate performance (mean across subjects).}
\label{tab:overall}
\small
\begin{tabular}{lccccccc}
\toprule
Split & Acc. & Macro-$F_{1}$ & $\kappa$ & Weighted $F_{1}$ & Bal. Acc. & MCC & ECE \\
\midrule
Validation & \textbf{61.7\%} & 36.8\% & 0.383 & 66.0\% & 39.1\% & 0.394 & 0.075 \\
Test       & \textbf{67.0\%} & 35.1\% & 0.394 & 70.1\% & 39.9\% & 0.410 & 0.081 \\
\bottomrule
\end{tabular}
\end{table}

\subsection{Error structure and calibration}
\begin{figure}[t]
  \centering
  \includegraphics[width=\linewidth]{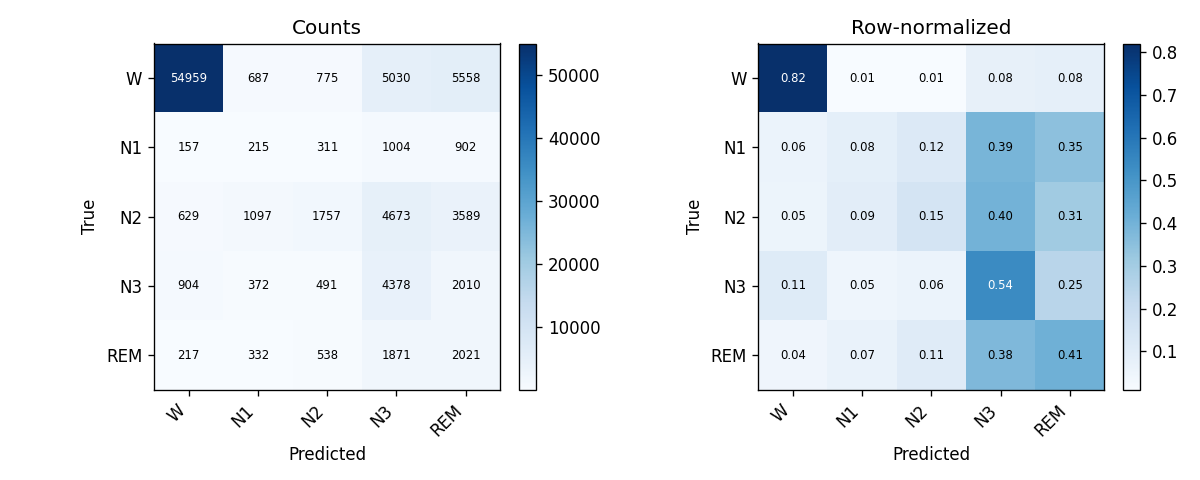}
  \caption{Test confusion matrix (counts and row-normalized). N1 remains hardest; W is easiest.}
  \label{fig:confmat-test}
\end{figure}

Normalized confusions on test (Fig.~\ref{fig:confmat-test}) show dominant errors around N1 and its neighbors (N2/REM), consistent with prior single-lead reports \citep{supratak2017deepsleepnet,Perslev2021USleep,eldele2021attnsleep}. Reliability is moderately under-confident at mid-range probabilities; ECE is $\sim$0.08 (Table~\ref{tab:overall}). Detailed calibration curves, stage distributions, and transition matrices are provided in the appendix.

\subsection{Calibration (added back)}
\begin{figure}[t]
  \centering
  \begin{minipage}[t]{0.48\linewidth}
    \centering
    \includegraphics[width=\linewidth]{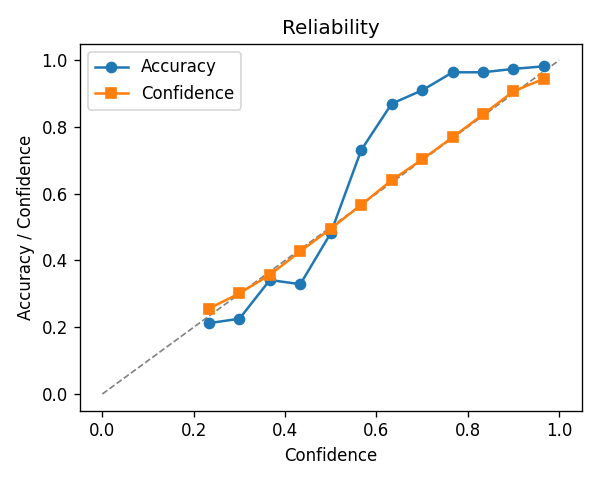}
  \end{minipage}\hfill
  \begin{minipage}[t]{0.48\linewidth}
    \centering
    \includegraphics[width=\linewidth]{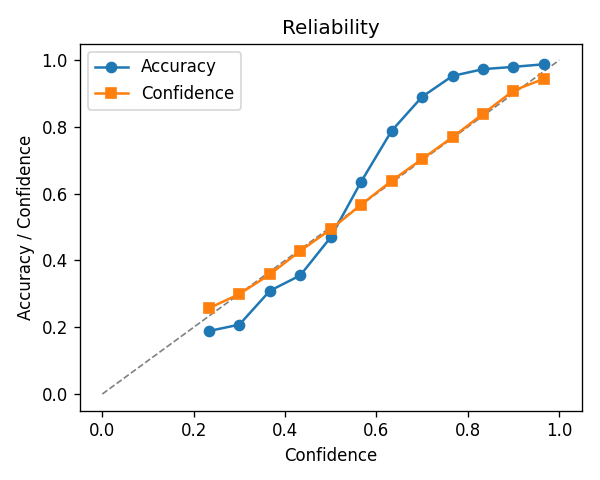}
  \end{minipage}
  \caption{Reliability diagrams for validation (left) and test (right).}
  \label{fig:reliability-both}
\end{figure}

\noindent\textit{Description.}
Reliability curves indicate mild under-confidence at mid-range probabilities and improved calibration at high confidence; the expected calibration error (ECE) is $\sim$0.08 on both splits (Table~\ref{tab:overall}), consistent with single-lead EEG staging reports \citep{supratak2017deepsleepnet,Perslev2021USleep,eldele2021attnsleep,guo2017calibration}.

\subsection{Stage distribution (added back)}
\begin{figure}[t]
  \centering
  \begin{minipage}[t]{0.48\linewidth}
    \centering
    \includegraphics[width=\linewidth]{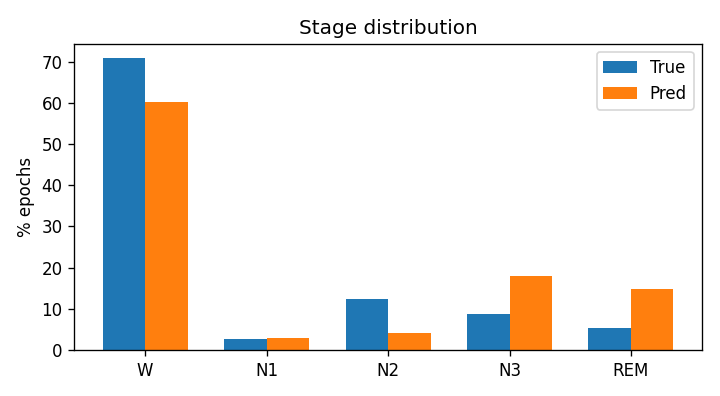}
  \end{minipage}\hfill
  \begin{minipage}[t]{0.48\linewidth}
    \centering
    \includegraphics[width=\linewidth]{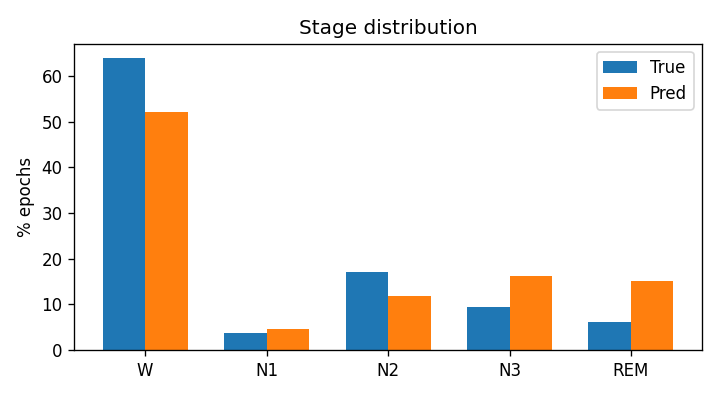}
  \end{minipage}
  \caption{Predicted vs.\ empirical stage distributions for test (left) and validation (right).}
  \label{fig:stage-dist-both}
\end{figure}

\noindent\textit{Description.}
Predicted hypnogram statistics track empirical distributions with expected deviations: under-prediction of N2 and slight over-prediction in N3/REM, mirroring confusion patterns (Fig.~\ref{fig:confmat-test}) and prior single-lead results \citep{supratak2017deepsleepnet,Perslev2021USleep,alvarezestevez2021interdb}.

\subsection{Ablations and safeguards (qualitative summary)}
BN-only improves over frozen inference by aligning statistics at deployment \citep{li2016adabn,Li2018AdaBN}. Tent adds further gains through entropy minimization with negligible extra compute \citep{wang2021tent}. The entropy gate suppresses updates on artefactual or low-information windows (near-uniform or spiky overconfidence), and the EMA reset curbs drift, echoing continual TTA stabilizers \citep{wang2022cotta}. We keep all adaptation hyperparameters fixed from validation when evaluating on test, and we never use test labels during adaptation. Subject-level variability and additional plots are in the appendix.

\section{Conclusions}
We presented a simple, streaming TTA recipe for sleep staging that combines BN refresh and entropy minimization with two stability rails. It improves agreement over a frozen baseline with seconds-level latency and minimal memory, requires no source data or target labels, and integrates naturally with standard MNE-based pipelines on Sleep-EDF \citep{goldberger2000physionet,gramfort2013mne,gramfort2014mne}.

\paragraph{Limitations and outlook.}
This study uses single-lead EEG on one benchmark corpus; broader validation across multimodal PSG and datasets (MASS, SHHS, ISRUC) \citep{oreilly2014mass,quan1997shhs,khalighi2016isruc}, stronger yet safe TTA variants \citep{wang2021tent,wang2022cotta}, uncertainty-aware deferral and calibration tuning \citep{guo2017calibration}, and edge profiling are promising next steps. We follow AASM conventions \citep{RechtschaffenKales1968,Iber2007AASM,Berry2012AASMUpdate,berry2015aasm} and keep all test-time updates label-free and streaming-compatible.%
\footnote{Configs and scripts for end-to-end reproduction will be released in an anonymized repository upon acceptance.}


\newpage
\section{References}
\nocite{*}
\bibliographystyle{plain}
\bibliography{StableSleep}

\newpage
\appendix
\section{Supplementary figures}

\begin{figure}[h]
  \centering
  \includegraphics[width=\linewidth]{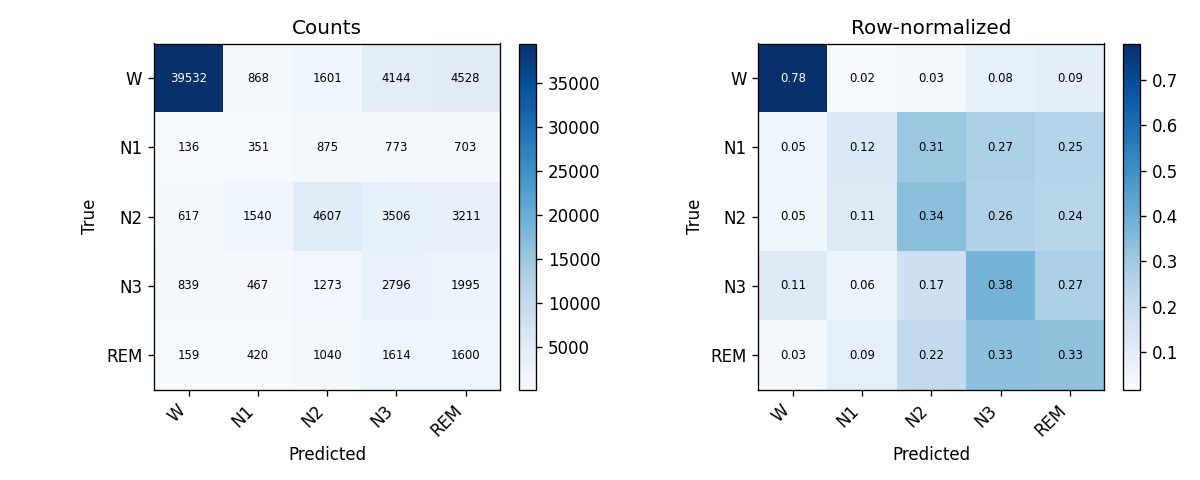}
  \caption{Validation confusion matrix (counts and row-normalized).}
  \label{fig:confmat-val}
\end{figure}

\begin{figure}[h]
  \centering
  \includegraphics[width=\linewidth]{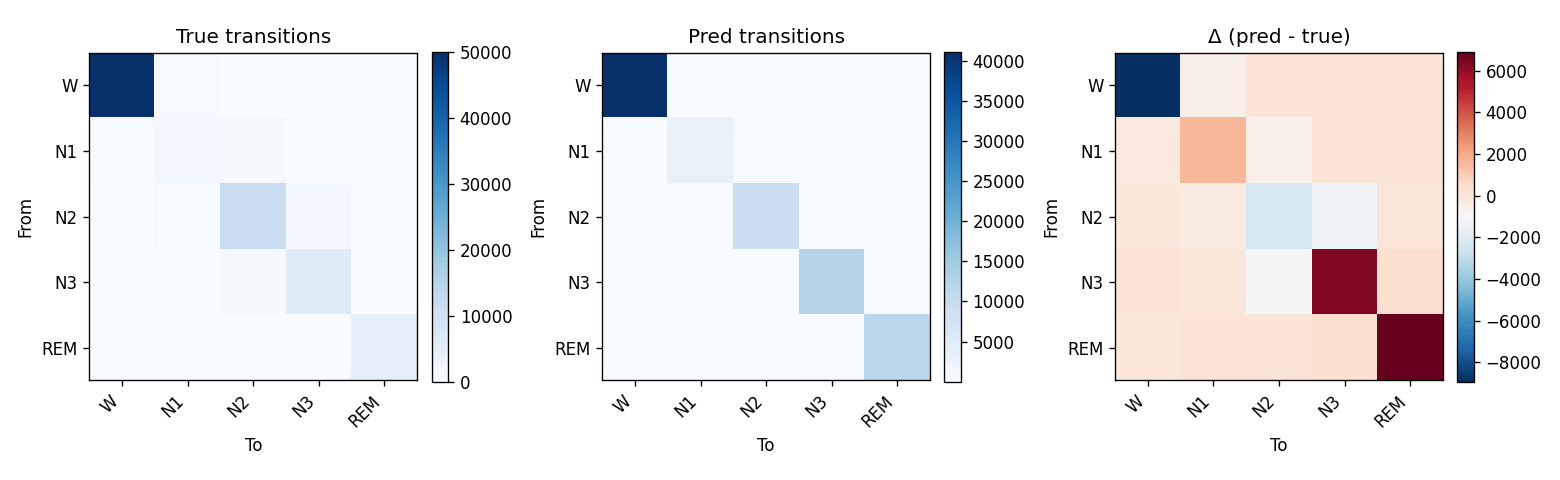}
  \caption{Stage transition matrices and residuals (validation).}
  \label{fig:transitions-val}
\end{figure}

\begin{figure}[h]
  \centering
  \includegraphics[width=\linewidth]{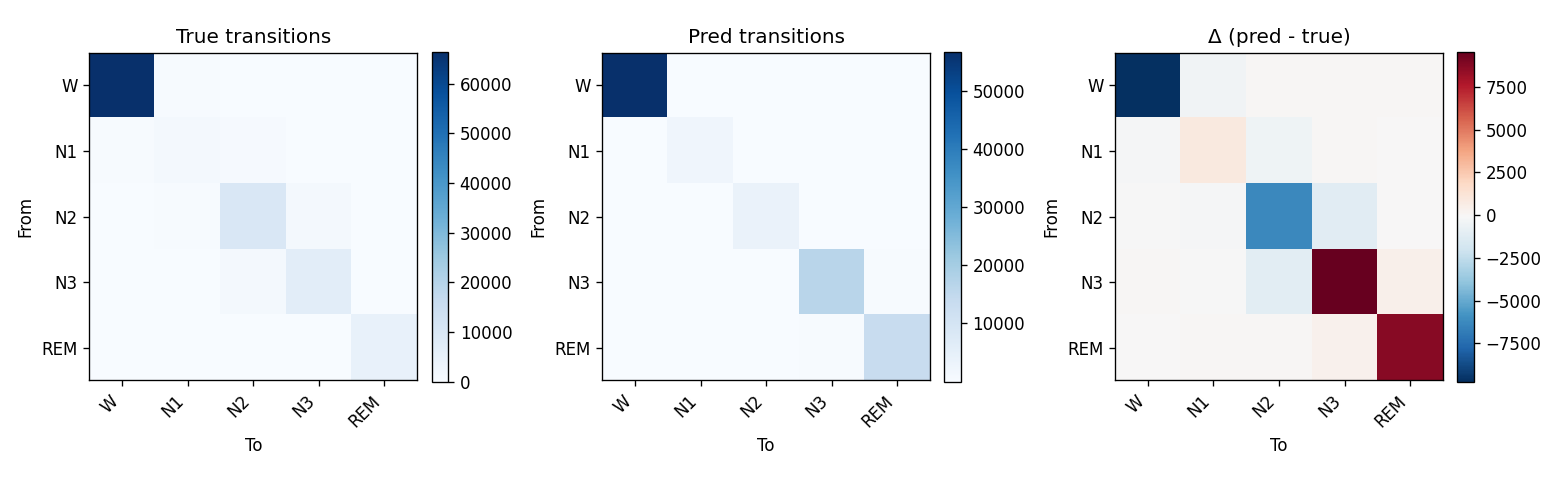}
  \caption{Stage transition matrices and residuals (test).}
  \label{fig:transitions-test}
\end{figure}

\begin{figure}[h]
  \centering
  \includegraphics[width=\linewidth]{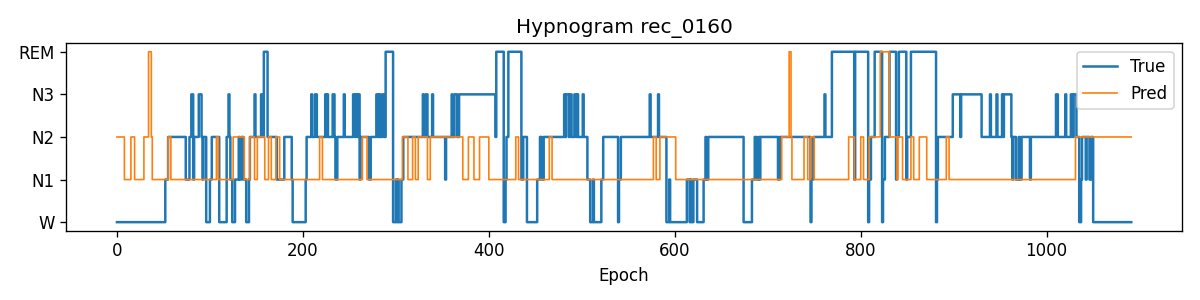}
  \caption{Example hypnogram from a test subject.}
  \label{fig:hypno-ex}
\end{figure}

\begin{figure}[h]
  \centering
  \includegraphics[width=0.48\linewidth]{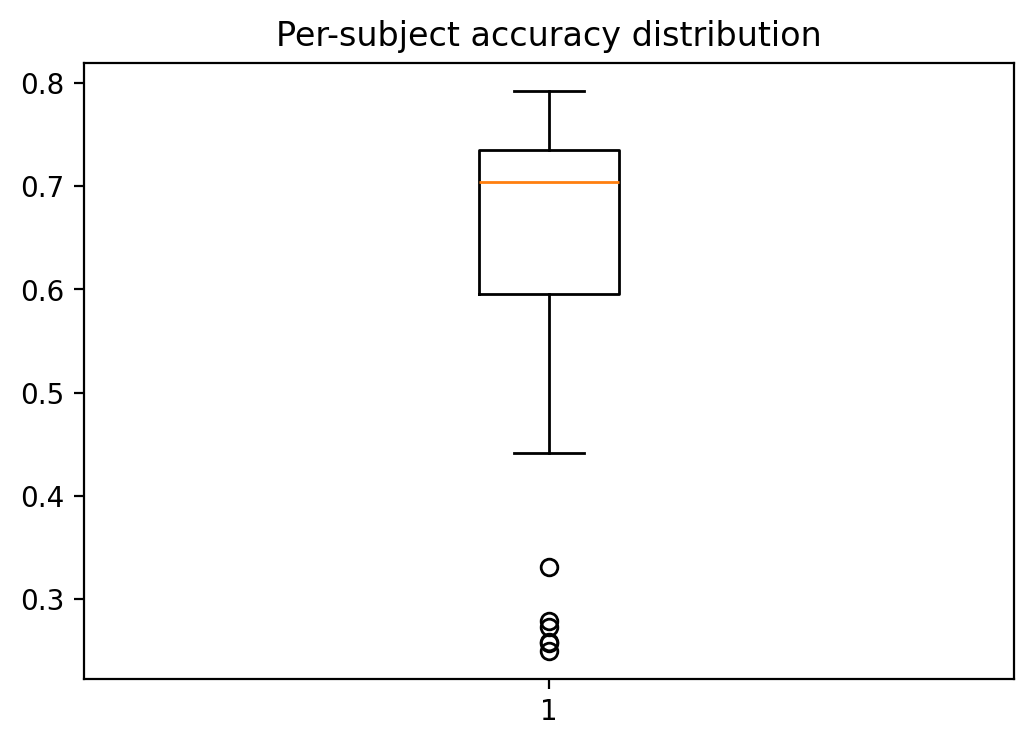}\hfill
  \includegraphics[width=0.48\linewidth]{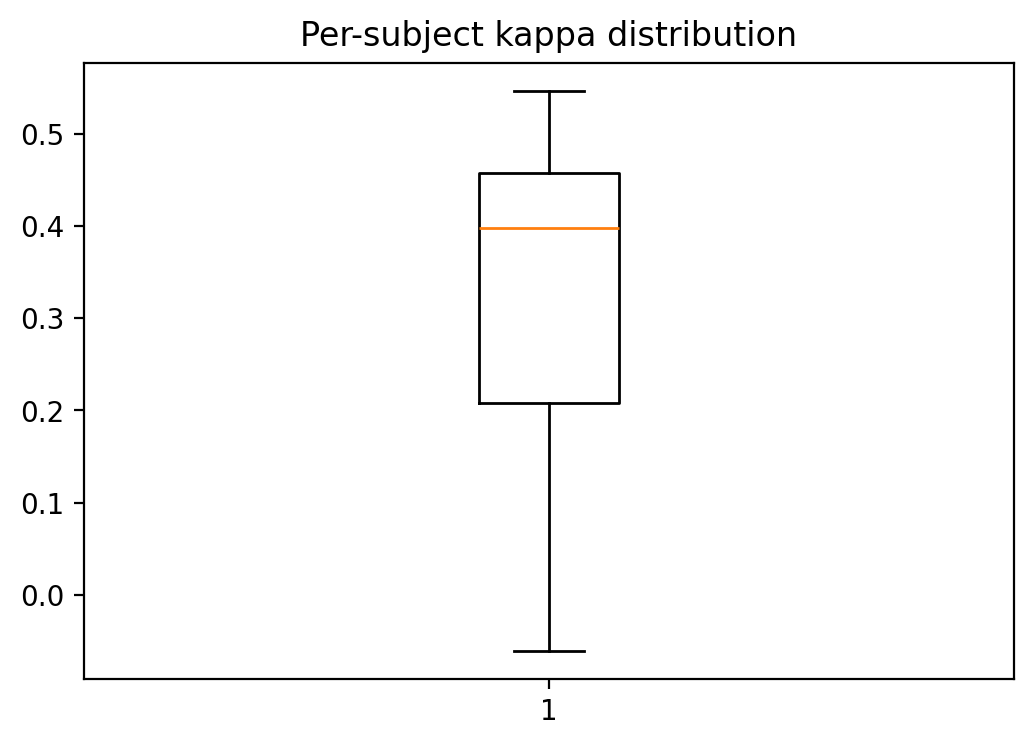}
  \caption{Subject-wise distributions (accuracy, $\kappa$).}
  \label{fig:subj-box}
\end{figure}

\FloatBarrier

\section{Reproducibility Details}
\label{app:repro}

\subsection{Compute \& Environment}
\label{app:compute}
\begin{itemize}
  \item \textbf{Hardware:} Apple MacBook Pro (M4 Pro; Apple Silicon, unified memory), NVMe SSD.
  \item \textbf{OS:} macOS (Apple Silicon build).
  \item \textbf{Acceleration:} PyTorch MPS backend (\texttt{torch.backends.mps}); CUDA not used.
  \item \textbf{Software:} Python~3.12; PyTorch~$\ge$2.2 (MPS); NumPy~$\ge$1.24; SciPy~$\ge$1.10; scikit-learn~$\ge$1.3; MNE~$\ge$1.4; Matplotlib~$\ge$3.7; tqdm~$\ge$4.65; PyYAML~$\ge$6.0.
  \item \textbf{Runtime:} Source training (37 epochs): ~30\,min/epoch, total 18\,h.
  \item \textbf{Determinism:} Fixed seeds for \texttt{torch} and \texttt{numpy}; subject-wise splits fixed.
\end{itemize}


\end{document}